\documentclass[lettersize,journal]{IEEEtran}
\usepackage{amsmath,amsfonts}
\usepackage{algorithmic}
\usepackage{algorithm}
\usepackage{array}
\usepackage[caption=false,font=normalsize,labelfont=sf,textfont=sf]{subfig}
\usepackage{textcomp}
\usepackage{stfloats}
\usepackage{url}
\usepackage{verbatim}
\usepackage{graphicx}
\usepackage{cite}
\usepackage{mathrsfs}
\usepackage{multirow}
\usepackage{makecell}
\usepackage{booktabs}
\usepackage{hyperref}

\begin{document}


\title{ERMV: Editing 4D Robotic Multi-view images\\to enhance embodied agents}

\author{Chang Nie, Guangming Wang, Zhe Liu,~\IEEEmembership{Member,~IEEE,} and Hesheng Wang,~\IEEEmembership{Senior Member,~IEEE,}
\thanks{This work was supported in part by the Natural Science Foundation of China under Grant 62225309, U24A20278, 62361166632 and U21A20480. (Corresponding Author: Hesheng Wang, e-mail: wanghesheng@sjtu.edu.cn)}%
\thanks{Chang Nie, Zhe Liu and Hesheng Wang are with School of Automation and Intelligent Sensing, Shanghai Jiao Tong University and Key Laboratory of System Control and Information Processing, Ministry of Education of China, Shanghai 200240, China.}
\thanks{Guangming Wang is with the Department of Engineering, Cambridge
University, Cambridge CB2 1TN, UK.}}

\markboth{Journal of \LaTeX\ Class Files,~Vol.~14, No.~8, August~2021}%
{Shell \MakeLowercase{\textit{et al.}}: A Sample Article Using IEEEtran.cls for IEEE Journals}


\maketitle

\begin{abstract}
    Robot imitation learning relies on 4D multi-view sequential images. However, the high cost of data collection and the scarcity of high-quality data severely constrain the generalization and application of embodied intelligence policies like Vision-Language-Action (VLA) models. Data augmentation is a powerful strategy to overcome data scarcity, but methods for editing 4D multi-view sequential images for manipulation tasks are currently lacking. Thus, we propose ERMV (Editing Robotic Multi-View 4D data), a novel data augmentation framework that efficiently edits an entire multi-view sequence based on single-frame editing and robot state conditions. This task presents three core challenges: (1) maintaining geometric and appearance consistency across dynamic views and long time horizons; (2) expanding the working window with low computational costs; and (3) ensuring the semantic integrity of critical objects like the robot arm. ERMV addresses these challenges through a series of innovations. First, to ensure spatio-temporal consistency in motion blur, we introduce a novel Epipolar Motion-Aware Attention (EMA-Attn) mechanism that learns pixel shift caused by movement before applying geometric constraints. Second, to maximize the editing working window, ERMV pioneers a Sparse Spatio-Temporal (STT) module, which decouples the temporal and spatial views and remodels a single-frame multi-view problem through sparse sampling of the views to reduce computational demands. Third, to alleviate error accumulation, we incorporate a feedback intervention Mechanism, which uses a Multimodal Large Language Model (MLLM) to check editing inconsistencies and request targeted expert guidance only when necessary. Extensive experiments demonstrate that ERMV-augmented data significantly boosts the robustness and generalization of VLA models in both simulated and real-world environments. Furthermore, ERMV can transform simulated images into a realistic style, effectively bridging the sim-to-real gap. The code will be available at \url{https://github.com/IRMVLab/ERMV}.
\end{abstract}


\section{Introduction}
\label{sec:intro}
\IEEEPARstart{I}{mitation} learning, which acquires skills by observing and mimicking expert demonstrations, has become a cornerstone for training embodied agents such as Vision-Language-Action (VLA) models. The core of this paradigm lies in learning a complex mapping from multi-view, temporal images (e.g. 4D sequences) to a trajectory of actions. However, the high cost and time-intensive nature of collecting high-quality expert demonstrations lead to a significant data bottleneck. This scarcity severely limits the generalization and robustness of VLAs in open-world scenarios. While data augmentation is a promising approach, existing methods, such as CACTI \cite{mandi2022cacti} and ROSIE \cite{Yu2023Scaling}, focus on editing single, static images. This is fundamentally insufficient for modern VLAs like RDT \cite{liu2024rdt} and OpenVLA \cite{kim2024openvla}, which demand spatio-temporally continuous 4D data for training. This discrepancy reveals a largely unexplored frontier in data augmentation: editing 4D robotic multi-view sequential images.

\begin{figure}[t]
  \centering
   \includegraphics[width=0.99\linewidth]{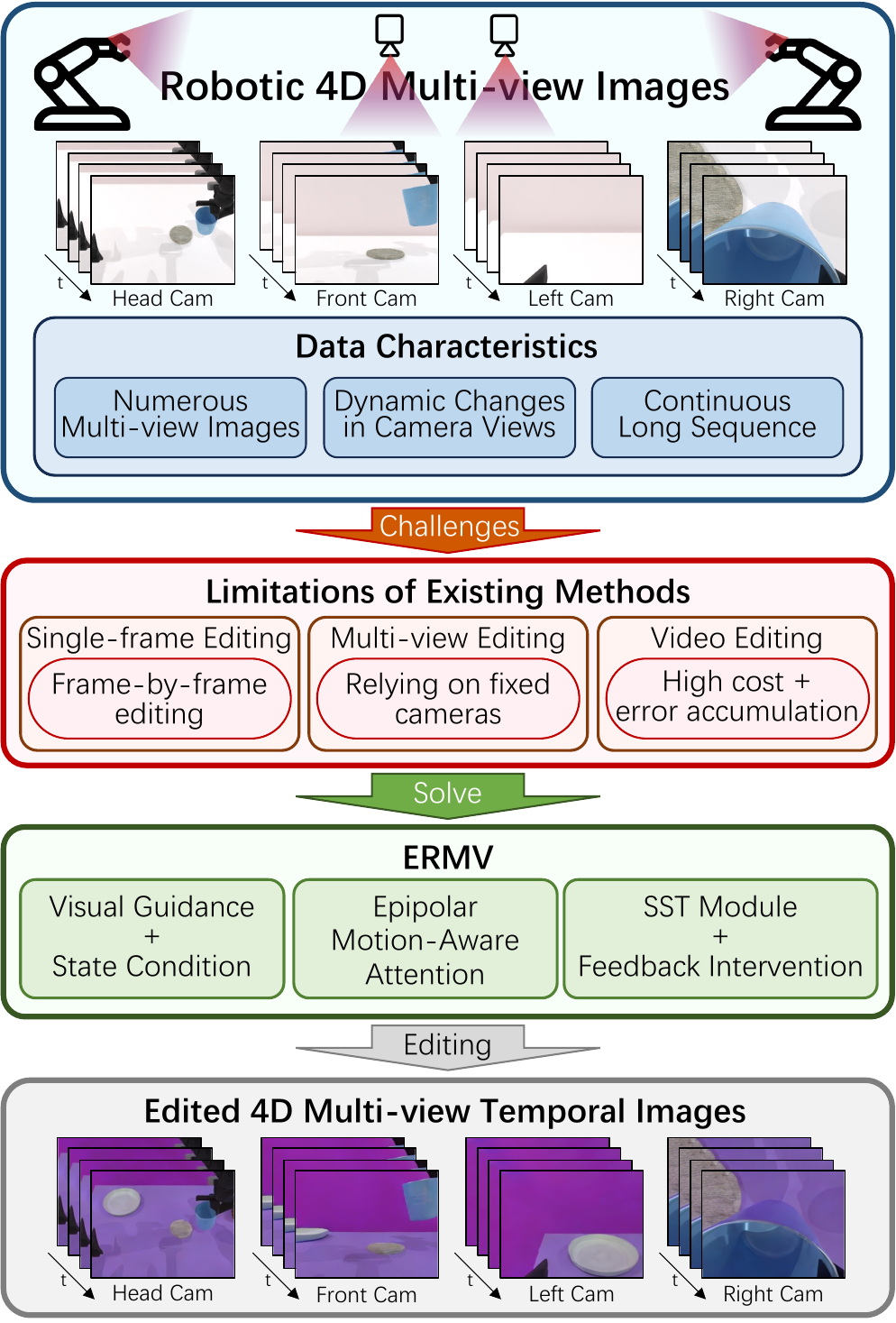}
    \vspace{-20pt}
   \caption{\textbf{The challenges of editing 4D robotic multi-view images.} ERMV incorporates a series of innovative approaches to address the challenges faced by existing methods, enabling accurate and convenient 4D sequence editing.}
   \vspace{-8pt}
   \label{fig:overall}
\end{figure}

The difficulty of editing 4D robotic multi-view sequential images stems from three fundamental technical obstacles: maintaining spatio-temporal consistency, operating within a small working window due to computing costs, and ensuring the quality of task-critical objects. First and foremost, it is essential to maintain the spatio-temporal consistency while editing 4D data. Spatially, existing methods focus on editing the fixed multi-cameras images on autonomous vehicles, maintaining spatial consistency through fixed adjacent view relationships \cite{gao2023magicdrive, yang2023bevcontrol, swerdlow2024street}. However, robotic manipulation involves a dynamically changing multi-camera system, rendering these fixed-cameras methods invalid. Temporally, edits must remain coherent across long time horizons. Current multi-image editing methods can only achieve sequential single-view video editing \cite{deng2024dragvideo, liu2024video, zeng2023visual, huang2025diffusion, yi2024feditnet}. Multi-view video editing has not been solved due to the difficulty of ensuring both temporal and spatial consistency. In addition, rebuilding 3D scenes and editing them can solve the problem of consistency across multiple viewpoints \cite{wu2024gaussctrl, kwon2025efficient}. But it is hard to accurately edit interactions between robots and objects. Moreover, a critical and often overlooked difficulty is the motion blur resulting from simultaneous camera and object movements. This dynamic effect breaks the assumptions of standard geometric constraints, such as epipolar lines, making it difficult to establish accurate feature correspondences. Therefore, existing methods that lack effective motion modeling struggle to restore motion blur and maintain realism.

Another formidable challenge is the small working window limited by computing costs and efficiency. State-of-the-art generative video models rely on dense spatio-temporal attention to establish temporal correlations. This means that extracting long temporal features with a large working window requires large GPU memory \cite{brooks2022generating, zhou2024storydiffusion, liu2024evalcrafter}. This hardware condition limits their accessibility and practical application. On the contrary, given that most robotic manipulation scenes involve gradual changes in a relatively static background, there are not many inter-frame distinctions that need to be captured by such a dense attention mechanism. Therefore, achieving a convenient and low-cost sequence editing framework without compromising 4D consistency is the key to improving the usability of generating or editing models in this domain. In addition, a single manipulation sequence can comprise thousands of images, making traditional view-by-view editing infeasible. Thus, an efficient and accurate approach to guide the editing is crucial.

The last challenge lies in the cumulative effect of errors. As edited images are autoregressively fed into the network as history frames, the accumulated errors gradually lead to a decline in image quality. Such an issue is particularly acute in robotic multi-view image editing tasks, which demand strict consistency of the robot arm and the manipulated object throughout the 4D sequence. This has become a common obstacle for existing methods in long-horizon data generation and editing \cite{you2024towards, ni2025maskgwm, li2024survey}. Consequently, establishing a strategy for effective evaluation and error correction is vital to ensuring the quality of the edited long sequence.

As shown in Fig. \ref{fig:overall}, to address these challenges, we propose ERMV (Editing Robotic Multi-View 4D data), a novel editing framework for enhance embodied agents. ERMV introduces a series of solutions to tackle the core challenges of 4D data editing. First, to avoid the ambiguity of text prompts, we employ a precise visual guidance strategy, where a single, user-edited image serves as a clear blueprint for desired changes.

Second, to expand the working window while retaining small computational costs, we pioneer a Sparse Spatio-Temporal (SST) module. By sparsely sampling views in a spatio-temporally decoupled large working window, ERMV remodels the video editing task as a low-cost, single-frame multi-view editing problem, allowing it to be trained on a single consumer GPU.

Third, to establish accurate geometric constraints and preserve motion blur in dynamic environments, we design a novel Epipolar Motion-Aware Attention (EMA-Attn) mechanism. This mechanism explicitly accounts for motion blur by learning to predict motion-induced pixel offsets before applying epipolar geometry to guide feature aggregation, ensuring robust correspondence during movements.

Finally, to prevent the gradual degradation of core objects, like the robot arm or manipulated objects, from autoregressive error accumulation, we introduce a pragmatic feedback intervention mechanism. This strategy uses a Multimodal Large Language Model (MLLM) to automatically check the consistency of core objects before and after editing. Experts are then involved only when necessary to provide segmentation masks of the core objects.

We validated ERMV on the public RoboTwin simulation benchmark, where the ERMV augmented data significantly boosted the success rate and generalization of the VLA models in unknown environments. Furthermore, experiments on the real-world RDT dataset and our real dual-arm robot platform demonstrate that ERMV can effectively edit and augment real-world data to improve downstream policies performance and robustness. Moreover, ERMV can even edit simulation data to match real-world appearances, thereby significantly narrowing the sim-to-real gap and reducing the dependency on high-fidelity physical simulations.

The main contributions of ERMV are as follows:
\begin{itemize}
	\item We propose ERMV, a novel framework for editing 4D robotic multi-view sequential images. It can effectively alleviate the data scarcity problem in VLA training and enhance the robustness of VLA models.
 
        \item ERMV ensures spatio-temporal consistency under motion blur and achieves a large working window through an epipolar motion-aware Attention mechanism and a sparse spatio-temporal module. Furthermore, ERMV introduces a practical feedback intervention mechanism that utilizes MLLM to safeguard the consistency of the core objects with minimal expert effort.
 
	\item We conducted extensive experiments on simulation environments, real-world, and real robot platforms. Moreover, we verified its data augmentation effect on downstream VLA policies. In addition, ERMV can not only employed as a world model, but also bridge the sim-to-real gap.
    
\end{itemize}

\begin{figure*}[t]
  \centering
   \includegraphics[width=0.99\linewidth]{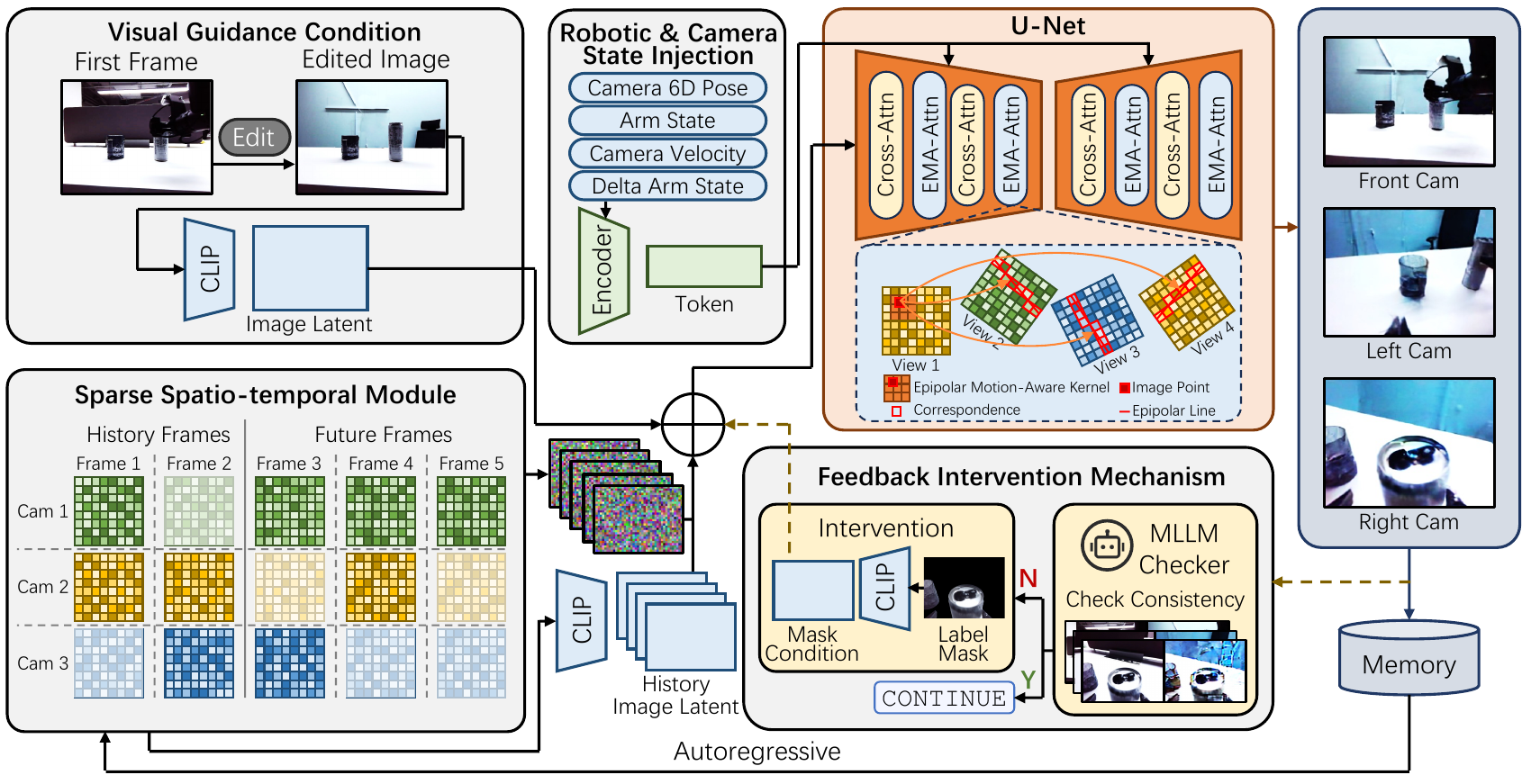}
    \vspace{-10pt}
   \caption{\textbf{The pipeline of the proposed ERMV.} \textbf{1.} Conditioning: The model takes a user-edited single frame as visual guidance and incorporates robotic \& camera state as physical conditions (Sec. \ref{visual_guidance}). \textbf{2.} Sparse Spatio-Temporal (SST) module: It employs sparse sampling to extract history frames and select the future frames to be edited in a large working window (Sec. \ref{state}). \textbf{3.} Diffusion model: Within the generative model, our proposed Epipolar Motion-Aware Attention (EMA-Attn) leverages epipolar geometry to guide feature fusion across multiple views and capture motion information (Sec. \ref{sstm}). (4) Feedback intervention mechanism: An intervention module powered by a Multi-modal Large Language Model (MLLM) assesses the quality of the key objects. If degradation is detected, it suggests expert intervention to inject a mask condition to ensure the quality of the core objects. (Sec. \ref{fpi}).}
   \vspace{-10pt}
   \label{fig:pipeline}
\end{figure*}

\section{Related Work}
\vspace{-4pt}
\label{sec:Related_Work}
\subsection{Robotics Images Generation and Editing}
\vspace{-4pt}
The emergence of high-fidelity generative and editing models, especially diffusion models, has opened up new frontiers in robotics. Current research in this domain utilizes these models in two main aspects: robotic high-level task planning and robotic training data augmentation.

\textbf{Generation for Task Planning.} Many studies use generative models to produce goal-oriented images to plan robot actions. Early works explored using pre-trained text-to-image models for zero-shot rearrangement of the final position of objects. For instance, DALL-E-Bot \cite{kapelyukh2023dall} first infers textual descriptions of objects in the scene. It then generates an image of the final state of the objects based on the desired goal. Finally, the robot is asked to place the objects in accordance with the generated image. Such a process realizes the planning of manipulating objects. Instead of generating a complete future video, SuSIE \cite{black2023zero} decomposes long-horizon tasks into more manageable key frames. SuSIE employs a hierarchical approach where a fine-tuned image-editing diffusion model acts as a high-level planner, proposing a future subgoal image. Then, a low-level, goal-conditioned policy is responsible for reaching that specific subgoal. More recently, the generative models have advanced toward creating comprehensive world models that function as interactive simulators for robot manipulation.
In summary, while these methods are useful for planning and generating images, their primary focus is on generating desired outcomes rather than editing existing 4D images for data augmentation.

\textbf{Editing for Data Augmentation.} 
Current robot imitation learning requires significant cost and human effort to collect high-quality data, which limits the robustness and generalization of models such as VLA. One promising direction to mitigate this is data augmentation, which aims to expand existing high-quality robotics datasets. Early methods use text-to-image generation models to add semantic diversity. Methods such as CACTI \cite{mandi2022cacti}, ROSIE \cite{Yu2023Scaling}, and GenAug \cite{chen2023genaug} demonstrate that applying inpainting techniques to single-view images could effectively modify scenes and diversify training data. For instance, CACTI \cite{mandi2022cacti} utilizes expert-collected data and augments it with scene and layout variations using generative models. ROSIE \cite{Yu2023Scaling} advances this by using text-to-image diffusion models to perform aggressive data augmentation, creating unseen objects, backgrounds, and distractors guided by text. GenAug \cite{chen2023genaug} introduces a framework for retargeting behaviors by generating semantically meaningful visual diversity in objects and backgrounds while aiming to maintain the functional invariance of actions. To achieve finer control and more physically plausible results, later work \cite{chen2024semantically} incorporate explicit 3D information, such as object meshes and depth guidance. Methods like RoboAgent \cite{bharadhwaj2024roboagent} sought to automate and scale this process further. RoboAgent integrates segmentation models like Segment Anything Model (SAM) \cite{kirillov2023segment} with inpainting to automatically identify and edit objects within a frame. However, a fundamental limitation plagues these approaches: they edit images frame-by-frame. This approach not only proves inefficient for video data but, more critically, fails to enforce the temporal and multi-view consistency essential for editing 4D robotic manipulation trajectories. While a more recent work, EVAC \cite{jiang2025enerverse}, a generative model and not an editing model, attempts to generate temporally coherent video conditioned on robot actions. But it learns consistency implicitly by merging multi-view inputs and relying on computationally intensive video models, rather than explicitly modeling the 3D geometry. This research highlights the challenges of editing robotic 4D data, which requires not only scalability and rich semantics but also guarantees spatio-temporal consistency.

\vspace{-10pt}
\subsection{Multi-view Images Generation and Editing}
\vspace{-4pt}
In addition to robotics, multi-view consistent generation techniques have been explored in areas such as autonomous driving and 3D object synthesis.

\textbf{Consistency in Structured Environments.} In autonomous driving, generating realistic and controllable data is crucial for robust simulation and model training. Several methods exploit the strong priors from fixed, circumferential camera rigs to synthesize street views from a unified Bird's-Eye-View (BEV) representation. This BEV space serves as a common ground for editing, allowing developers to craft specific scenarios. BEVGen \cite{swerdlow2024street} introduces a conditional generative model that synthesizes surrounding-view images from a semantic BEV layout. It utilizes an autoregressive transformer architecture and incorporates a novel pairwise camera bias, which learns the spatial relationship between different camera views to ensure their consistency. BEVControl \cite{yang2023bevcontrol} is proposed to achieve more accurate and finer-grained control over individual street-view elements. Instead of detailed semantic maps, BEVControl supports more flexible BEV sketch layouts that are easier for users to edit. It employs a two-stage, diffusion-based method featuring a ``Controller" for geometric consistency and a ``Coordinator" with a cross-view-cross-element attention mechanism to maintain appearance consistency across the different viewpoints. More recently, MagicDrive \cite{gao2023magicdrive} achieves the state-of-the-art by enabling diverse and direct 3D geometry control. It addresses the limitations of BEV-only conditioning, which can lead to geometric ambiguities such as incorrect object heights or road elevations. MagicDrive uses a diffusion model to separately encode various inputs, including BEV roadmaps, explicit 3D bounding boxes, camera poses, and text descriptions. Its multi-view consistency is achieved through a cross-view attention with hard-coded neighboring views. These existing multi-view editing methods rely heavily on the fixed relative positions of multiple cameras. However, these methods cannot solve the problem of editing dynamically changing multi-view images during robot manipulation.

\textbf{3D Asset Generation and Editing.} In 3D asset-related domains, many methods enforce multi-view consistency through geometric constraints or 3D representations. Foundational work in this area, like Zero-1-to-3 \cite{liu2023zero}, demonstrated that pre-trained 2D diffusion models could be fine-tuned to understand relative camera transformations. This model can then use the learned geometric prior to zero-shot synthesize novel views from a single image. Building on this, feed-forward frameworks like InstantMesh \cite{xu2024instantmesh} achieve significant efficiency by first using a multi-view diffusion model to generate a sparse set of consistent images, which are then fed into a Large Reconstruction Model (LRM) to directly produce a high-quality 3D mesh in seconds. To further enhance geometric coherence, 3D-Adapter \cite{chen20243d} introduces a plug-in module that injects explicit 3D awareness into the denoising process. It operates via a ``3D feedback augmentation" loop, where intermediate multi-view features are decoded into a 3D representation, such as 3D Gaussian Splatting (3DGS). Moreover, in the domain of 3D editing, DGE \cite{chen2024dge} skips slow iterative optimization by editing 2D images with multi-view consistency. Its spatio-temporal attention and epipolar constraints are extracted from the scene geometry to augment the 2D editing, allowing for a direct and efficient update to the 3DGS model. For the complex task of 3D inpainting in unconstrained scenes, IMFine \cite{shi2025imfine} proposes a geometry-guided pipeline that uses test-time adaptation to fine-tune a multi-view refinement network for each scene, correcting artifacts from warping an inpainted reference view to others. However, these methods lack mechanisms for handling motion blur images and complex tool-object interactions inherent in robot manipulation tasks.

In summary, existing research fails to address a key need in robotics: a method for consistent and controlled editing of multi-view temporal images of dynamic manipulation tasks. Our work aims to fill this gap by proposing a framework that explicitly models spatio-temporal consistency and allows for easy editing of 4D robotics data.

\vspace{-7pt}
\section{Method}
\vspace{-4pt}

\subsection{Problem Formulation and Framework Overview}
\textbf{Problem Formulation.} Given a 4D robotic manipulation trajectory $\mathcal{T} = (\mathbf{X}_t, \mathbf{a}_t), t=1 \cdots T$, where $\mathbf{X}_t=x_{t}^{(v)}, v=1 \cdots N$ represents the set of $N$ multi-view images at timestep $t$, and $\mathbf{a}_t \in \mathcal{A}$ is the corresponding robot action. The primary objective is to perform targeted edits on the image sequence $\mathcal{X}=\mathbf{X}_t, t=1 \cdots T$ to generate a new, visually distinct yet semantically consistent sequence $\mathcal{X}^{\prime }$. This new sequence, when paired with the original, unmodified action sequence $\left \{ \mathbf{a}_t, t=1 \cdots T \right \} $, forms an augmented data pair $\mathcal{T}^{\prime } = (\mathbf{X}_t^{\prime }, \mathbf{a}_t), t=1 \cdots T$. This process serves as a powerful data augmentation strategy to alleviate the data scarcity problem in embodied intelligence.

\textbf{Framework Overview.} To achieve controllable editing, we propose ERMV (Editing for Robotic Multi-view data), a framework built upon the principles of Latent Diffusion Models (LDMs) \cite{rombach2022high}. The core of our method is a conditional generator $G_\theta $, which synthesizes the edited multi-view sequence $\mathcal{X}^{\prime }$ based on the original images $\mathcal{X}$, a fine-grained visual guidance signal $\mathcal{C}_{guide}$, robot-centric state information $\mathcal{C}_{state}$, and memory features $\mathcal{C}_{history}$. The overall generation process can be formulated as learning a conditional probability distribution:
\begin{equation}
p(\mathcal{X}' | \mathcal{X}, \mathcal{C}_{guide}, \mathcal{C}_{state}, \mathcal{C}_{history})
\end{equation}

Our framework, depicted in Fig. \ref{fig:pipeline}, systematically overcomes the core difficulties of this task. The process begins by establishing precise visual guidance (Section \ref{visual_guidance}). To overcome the ambiguity of text prompts, we use a single, edited image as a rich visual blueprint for the desired modifications. For consistent editing across views and timesteps, we ground the model in the physical reality of the scene through spatio-temporal attentions that explicitly inject camera poses, robot states, and their temporal dynamics (Section \ref{state}). Furthermore, ERMV maximizes the working window under limited conditions by a Sparse Spatio-Temporal (SST) module (Section \ref{sstm}), a strategy that captures long-range memory without prohibitive computational costs by reframing video generation as a single-frame, multi-image problem. Within the generation model, ERMV introduces an Epipolar Motion-Aware attention (EMA-Attn) to capture motion features (Section \ref{epipolar}), realistically rendering the motion blur common in robotic manipulation. Finally, to prevent semantic drift and error accumulation, a feedback intervention mechanism leverages an MLLM to safeguard the integrity of critical scene elements like the robot arm and the manipulated object (Section \ref{fpi}). Then, experts are asked to correct errors only when necessary. The images edited in the current working window are stored in memory to autoregressively edit future frames.

The diffusion process operates in a latent space for computational efficiency, using a pre-trained autoencoder with an encoder $\mathbf{E}$ and a decoder $\mathbf{D}$ in the generator $G_\theta $. The forward process adds Gaussian noise $\epsilon$ to the latent representations $\mathbf{z}_0=\mathbf{E}(\mathcal{X})$ to produce noisy latents $\mathbf{z}_t$. The model $G_\theta $ is trained to predict the added noise from $\mathbf{z}_t$, conditioned on the timestep $t$ and our comprehensive set of conditions $\mathcal{C}=\left \{ \mathcal{C}_{guide},\mathcal{C}_{state}, \mathcal{C}_{history} \right \} $. The loss function is:
 \begin{equation}
\mathcal{L}_{\text{LDM}} = \mathbb{E}_{\mathbf{E}(\mathcal{X}), t, \mathcal{C}, \epsilon} \left[ \left\| \epsilon - G_\theta (\mathbf{z}_t, t, \mathcal{C}) \right\|^2 \right]
\label{eq:loss}
\end{equation}

\subsection{Visual Guidance Condition}
\label{visual_guidance}
A fundamental challenge in editing robot images is to follow expectations accurately. While text prompts are standard in creative image editing \cite{wu2024towards,chiu2024brush2prompt, sharma2024sketch}, they fail to provide the granular geometric and spatial control essential for physically-grounded scenes. For example, a cue such as ``change background to office” lacks specificity to accurately define color, type, or orientation. The result may even conflict with robot actions. Consistency between edited images and actions is essential for training robust robot strategies.

Accurately editing a global image in advance to achieve the desired effect can effectively prevent misunderstandings of the desired edit. Thus, ERMV adopts a visual guidance strategy. We first select a globally-informative frame, typically the first frame from the primary camera, $x_{t=1}^{(v=1)}$, which captures the overall scene context. This frame is then meticulously edited, using either advanced inpainting models \cite{chiu2024brush2prompt, sharma2024sketch, he2024few} or manual editing, to create a target guidance image $x_{guide}^\prime$. This image serves as an explicit, unambiguous visual blueprint of the desired modifications. The guidance condition $C_{guide}$ is then processed by encoding this image using a pre-trained vision encoder, such as CLIP \cite{radford2021learning}:
\begin{equation}
\mathcal{C}_{guide} = \mathbf{E}_{\text{CLIP}}(x'_{guide})
\end{equation}
This rich embedding provides a precise, spatially-aware semantic target, enabling the diffusion model to propagate the edit consistently across all views and timesteps.

\subsection{Robotic and Camera State Injection}
\label{state}
Generating a coherent 4D sequence requires more than just a visual target. The model must understand the precise geometric and dynamic state of the robot and cameras at every moment. Lacking this information prevents correct positioning of the robot arm in each view and hinders the realistic rendering of motion blur. To accurately render the scene from robot camera viewpoints and timesteps, we inject explicit state information as part of the condition $C_{state}$, which consists of two components:

\textbf{Pose and State Conditioning.} For each target image $x_{t}^{(v)}$, we provide its corresponding camera pose $\mathbf{P}_t^{(v)}\in SE(3)$ and the robot action $q_t\in \mathbb{R} ^d$ (e.g., joint positions), where $d$ is the degree of freedom. This allows the model to ground the generation in the correct geometric context.

\textbf{Motion Dynamics Conditioning.} A common and challenging characteristic of robotic manipulation images is motion blur, caused by the simultaneous movement of the camera and objects. Failing to model this phenomenon will lead to unnaturally sharp and unrealistic videos. To explicitly capture these dynamics, we compute the temporal deltas of poses and states: camera motion $\Delta \mathbf{P}_t^{(v)}=\mathbf{P}_{t}^{(v)}-\mathbf{P}_{t-1}^{(v)}$, and robot motion $\Delta q_t=q_t-q_{t-1}$.

These static and dynamic features are concatenated to form a comprehensive state vector for each image: $\mathbf{c}_{t,v}=\left [ \mathbf{P}_{t}^{(v)},q_t,\Delta \mathbf{P}_t^{(v)},\Delta q_t \right ] $. This vector is then projected and encoded using a Multi-Layer Perceptron (MLP) with positional encoding $\Psi $ into a sequence of embedding tokens, which are fed into the cross-attention layers of the U-Net backbone:
\begin{equation}
\mathcal{C}_{state}^{(t,v)} = \Psi(\text{MLP}(\mathbf{c}_{t,v})).
\end{equation}

\begin{figure}[t]
  \centering
   \includegraphics[width=0.99\linewidth]{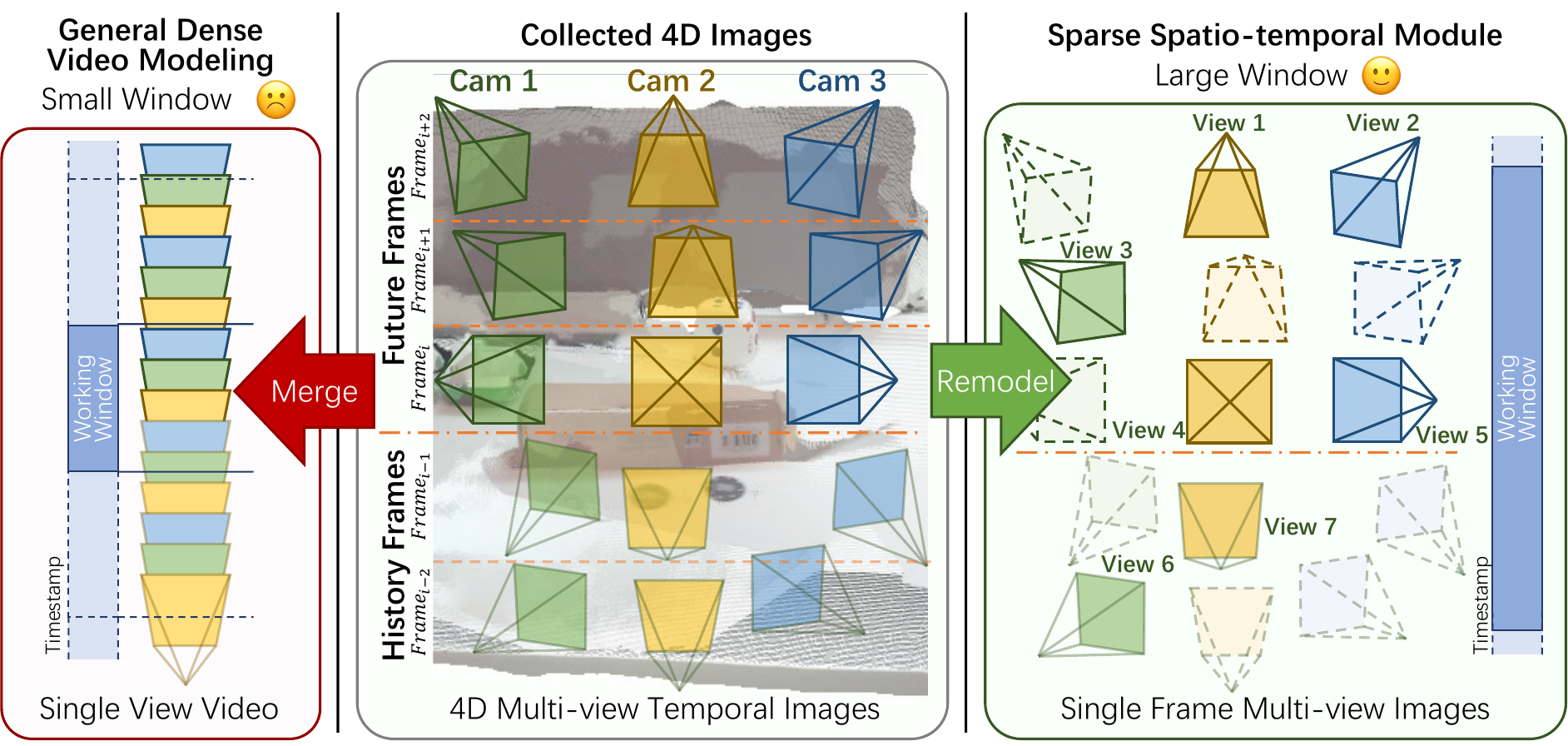}
    \vspace{-20pt}
   \caption{\textbf{Comparative Effects of Sparse Spatio-Temporal Module.} General methods (left) merge 4D multi-view temporal images into a single-view video, employing dense video techniques for spatio-temporal sequence. This results in working in small windows under limited GPU memory. We propose a SST module (right) that randomly samples views and reconstructs them into a single-frame multi-view problem, significantly reducing computational demands and expanding the working window. Integrated with geometry-guided EMA-Attn (see Sec. \ref{epipolar}), our approach accurately ensures spatio-temporal consistency.}

   \vspace{-8pt}
   \label{fig:SSTM}
\end{figure}

\subsection{Sparse Spatio-Temporal Module} 
\label{sstm}
Previous methods generally use video diffusion models \cite{ho2022video} to process multi-view temporal images \cite{huang2025enerverse, jiang2025enerverse}. This kind of model implicitly extracts geometric information through dense frame-by-frame cross-attention, leading to prohibitive computational costs, especially for a large working window. However, in many manipulation scenarios, the background is largely static and the images change slowly. Motivated by this observation, we propose a Sparse Spatio-Temporal (SST) module to maximize the working window within limited GPU memory.

As indicated in Fig. \ref{fig:SSTM}, given a sliding window of $L$ consecutive timesteps, instead of processing all $L\times N$ images, we randomly sample a fixed-size subset of $K$ images, where $K\ll L\times N$.  Let the sampled set be $\mathcal{X}_{sample}=\widetilde{x}_k,k=1\cdots K$, which includes history views $C_{history}$ and future views. Each sampled image $\widetilde{x}_k$ corresponds to an original image $x_{t_k}^{(v_k)}$ from timestep $t_k$ and view $v_k$. To preserve the original spatio-temporal structure lost during sampling, we explicitly encode the original indices $\left ( t_k, v_k \right ) $ and inject them as part of the condition for each respective image. Notably, ERMV not only injects history frames into the network as conditions, but also generates them together with future frames. This simultaneous generation approach allows future frames to better extract geometric structure information from history frames, thereby improving temporal consistency. By modeling the joint probability distribution:
\begin{equation}
p(\mathcal{X}_{sample}|\left \{ c_k \right \}_{k=1}^K),
\end{equation}
the model learns the feature of the entire sparse set of frames. Thus, the SST module allows the model to reason about a much wider temporal context with a fixed computational budget, effectively reframing the video generation problem as a low-cost, single-frame multi-view generation task.

\subsection{Epipolar Motion-Aware Attention}
\label{epipolar}
While sparse sampling is low-cost, it poses a new challenge: how to effectively propagate information and enforce geometric consistency between sparsely selected frames. Epipolar-guided attention \cite{huang2024epidiff} provides a strong geometric foundation. However, standard implementations fail to account for the motion blur in the robotic domain, as features sampled along precise epipolar lines in blurry images may not correspond to the true pixel positions.

\begin{figure}[t]
  \centering
   \includegraphics[width=0.7\linewidth]{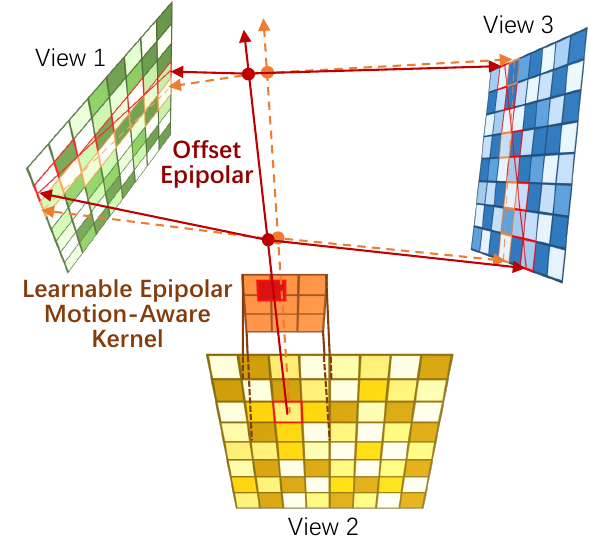}
    \vspace{-10pt}
   \caption{\textbf{Illustration of the Learnable Epipolar Motion-Aware Kernel.} To handle motion blur from camera and object movements, we propose a learnable motion-aware kernel. Instead of relying on rigid epipolar constraints (dashed orange lines), our method adaptively learns offset epipolar paths (solid red lines). This enables more robust cross-view feature correspondence in dynamic scenes by explicitly modeling motion-induced displacements.}

   \vspace{-8pt}
   \label{fig:kernel}
\end{figure}

To address this challenge, we introduce a novel Epipolar Motion-Aware (EMA) Attention. As shown in Fig. \ref{fig:kernel}, for a query pixel $\mathbf{p}_i$ in view $v_i$, ERMV does not assume its correspondence lies exactly on the epipolar line $\mathbf{l}_j=\mathbf{F}_{ij}\mathbf{p}_i$ in another view $v_j$. Instead, ERMV first predicts a motion-induced offset $\Delta \mathbf{p}_i $ using a small network $f_{blur}$ that considers local image features $\phi (\mathbf{p}_i) $ and the motion condition $\mathcal{C}_{state}^{(t_i,v_i)}$:
\begin{equation}
\Delta \mathbf{p}_i = f_{blur}(\phi(\mathbf{p}_i), \mathcal{C}_{state}^{(t_i,v_i)})
\end{equation}

\begin{figure}[t]
  \centering
   \includegraphics[width=0.99\linewidth]{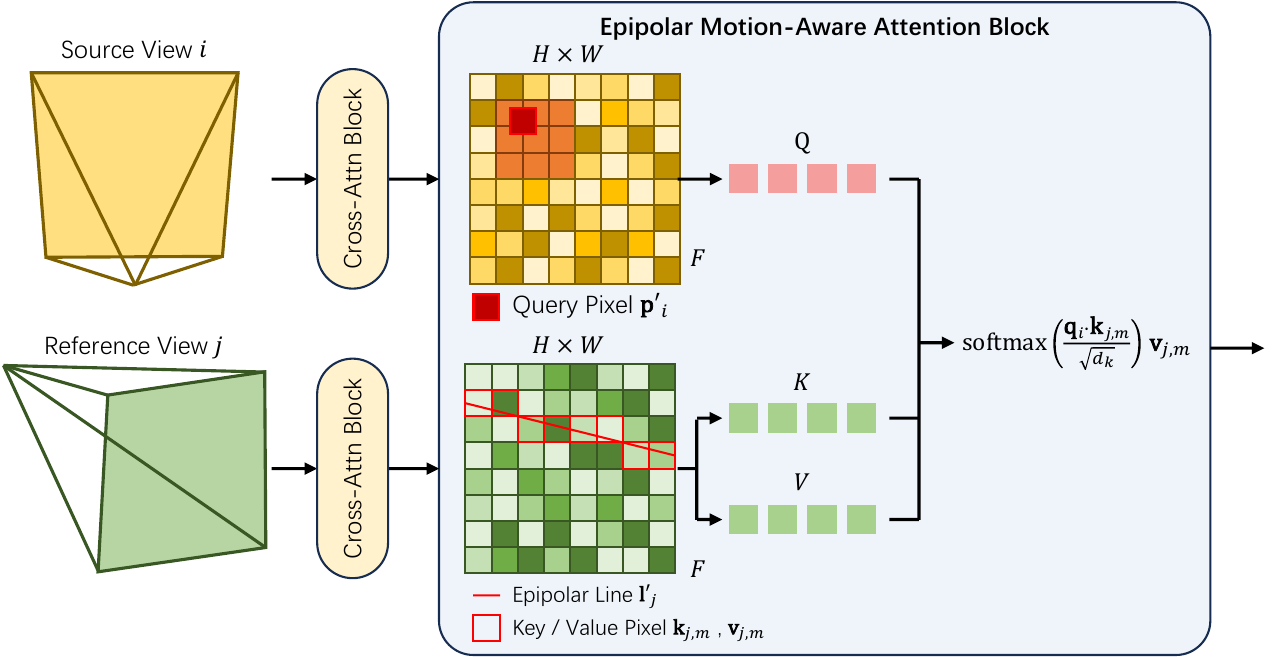}
    \vspace{-10pt}
   \caption{\textbf{Multi-View Feature Aggregation via EMA-Attn.} This block projects a query pixel from the source view to a shifted epipolar line in the target view. The attention mechanism is then constrained to the key/value pixels along this line, enabling efficient aggregation of multi-view features under geometric constraints to capture underlying motion.}

   \vspace{-8pt}
   \label{fig:attention}
\end{figure}

The feature aggregation is then performed along the shifted epipolar line corresponding to the new point $\mathbf{p}_i^\prime =\mathbf{p}_i+\Delta \mathbf{p}_i$. As illustrated in Fig. \ref{fig:attention}, the attention mechanism aggregates features from points $\mathbf{p}_{j,m}^\prime, m=1\cdots M$ sampled along the modified epipolar line $\mathbf{l}_j^\prime =\mathbf{F}_{ij}\mathbf{p}_i^\prime $:
\begin{equation}
\text{Attention}_{\text{EMA}}(\mathbf{q}_i, \mathbf{K}_j, \mathbf{V}_j) = \sum_{m=1}^{M} \text{softmax}\left(\frac{\mathbf{q}_i \cdot \mathbf{k}_{j,m}}{\sqrt{d_k}}\right) \mathbf{v}_{j,m}
\end{equation}
where $\mathbf{q}_i$ is the query feature at $\mathbf{p}_i^\prime$, and $\mathbf{k}_{j,m}, \mathbf{v}_{j,m}$ are key/value pairs at sampled points on the motion-aware epipolar line in view $v_j$. This allows the model to learn motion-specific correspondences, improving geometric consistency and realism.

\subsection{Feedback Intervention Mechanism}
\label{fpi}
Autoregressive image generation is prone to error accumulation \cite{sun2025ar, xie2025progressive, yin2025slow}, which can degrade quality and cause deviations from expectations. Moreover, the image quality of the manipulated objects and the robot arm is particularly important when training the VLA model. The degradation in these critical areas not only leads to visual inaccuracies, but also makes policy learning data invalid. Thus, it is crucial to preserve their quality.

\begin{table}[t]\small 
\centering
\caption{\textbf{Example chain-of-thought prompts for object consistency checking.}}
\vspace{-6pt}
\setlength{\tabcolsep}{1.0mm}
\renewcommand\arraystretch{1.1}
\begin{tabular}{c}
\toprule
\begin{tabular}[c]{@{}l@{}}
The image on the left is the original image and the image on the\\right is the image with the background edited. Do the \textit{\textless objects\textgreater}\\in the edited image match the original? If the degree of\\degradation of the objects is scored from 0-10, please rate the\\degree of degradation.\\
\textbf{Step 1}: Observe only the \textit{\textless objects\textgreater} in both images; the other\\backgrounds need no attention.\\ 
\textbf{Step 2}: If the \textit{\textless objects\textgreater} are not found in the image on the right,\\the image is severely degraded and can be scored directly at 10.\\ 
\textbf{Step 3}: Compare the similarity of the \textit{\textless objects\textgreater} in the two\\images and then score the degradation of the image.\\ 
\textbf{Step 4}: If the score is more than 5, it means serious degradation,\\output \verb|{"is consistent":False}| in JSON format,\\otherwise it means the degradation is not serious, output\\\verb|{"is consistent":True}|.\end{tabular}\\ \bottomrule
\end{tabular}
\label{tab:cot}
\vspace{-0.4cm}
\end{table}

\begin{figure}[t]
  \centering
   \includegraphics[width=0.99\linewidth]{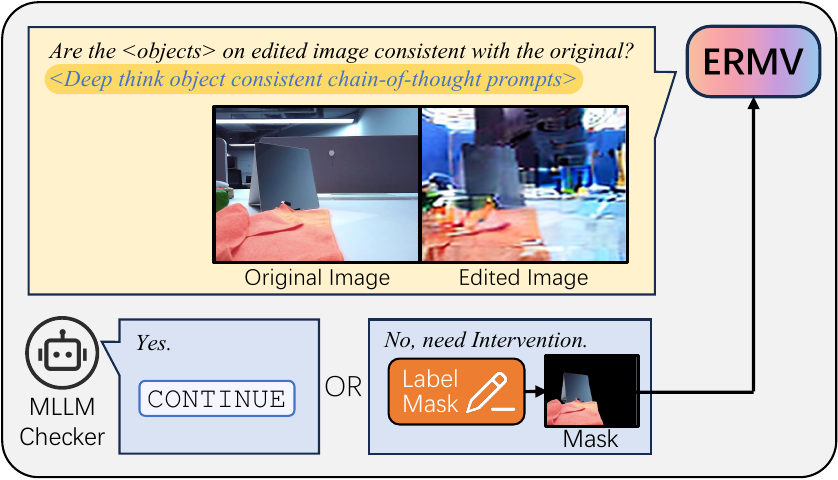}
    \vspace{-20pt}
   \caption{\textbf{MLLM-driven Feedback Correction for Editing Consistency.} To prevent content drift during iterative image editing, ERMV introduces a feedback intervention mechanism. A Multimodal Large Language Model (MLLM) is used as a consistency checker. When the MLLM identifies a significant deviation of the core objects in the edited image from the original, the system suggests the expert to label the core objects to be preserved with a mask. This mask is then encoded as a constraint to eliminate editing errors and restore object quality.}

   \vspace{-8pt}
   \label{fig:hil}
\end{figure}

A vanilla solution would be to segment the core objects, like robot arm and manipulated objects, in every frame to enforce their preservation. This solution can be implemented in two ways: on the one hand, training a universal segmentation model for so-called ``manipulated objects". However, the ``manipulated objects" are diverse and often novel. Furthermore, many robotic camera views are challenging and ego-centric. These obstacles make such training technically unfeasible. On the other hand, manually labeling the core objects can achieve great results. But the thousands of images that need to be labeled are excessively labor-intensive.

To solve this dilemma, we propose a feedback intervention mechanism. For each generated image $x_{t,(k)}^\prime $ at step k, we employ an MLLM $\Phi$ as an automated checker. It compares the generated image to the original $x_{t}$ with a task description prompt $\mathcal{P}_{CoT}$ based on the Chain-of-Thought (CoT) to check the consistency of the critical objects:
\begin{equation}
\text{is\_consistent} = \Phi ( x_t, x_{t,(k)}^\prime , \mathcal{P}_{CoT})
\end{equation}

The example prompt $\mathcal{P}_{CoT}$ is shown in the TABLE \ref{tab:cot}. If $\text{is\_consistent}$ is false, the system flags the image and suggests the expert to provide a segmentation mask $\mathbf{M}_t$ for the core objects in $x_{t,(k)}^\prime$. This mask is then incorporated as an additional condition $\mathcal{C}_{mask}$ for a corrective regeneration step. The advantage of this feedback loop is that it effectively prevents semantic drift with surgical precision while minimizing the expert annotation burden to only the few cases where the model falters. This feedback ensures the integrity of our augmented data without creating an unmanageable workflow.

\begin{figure*}[t]
  \centering
   \includegraphics[width=0.99\linewidth]{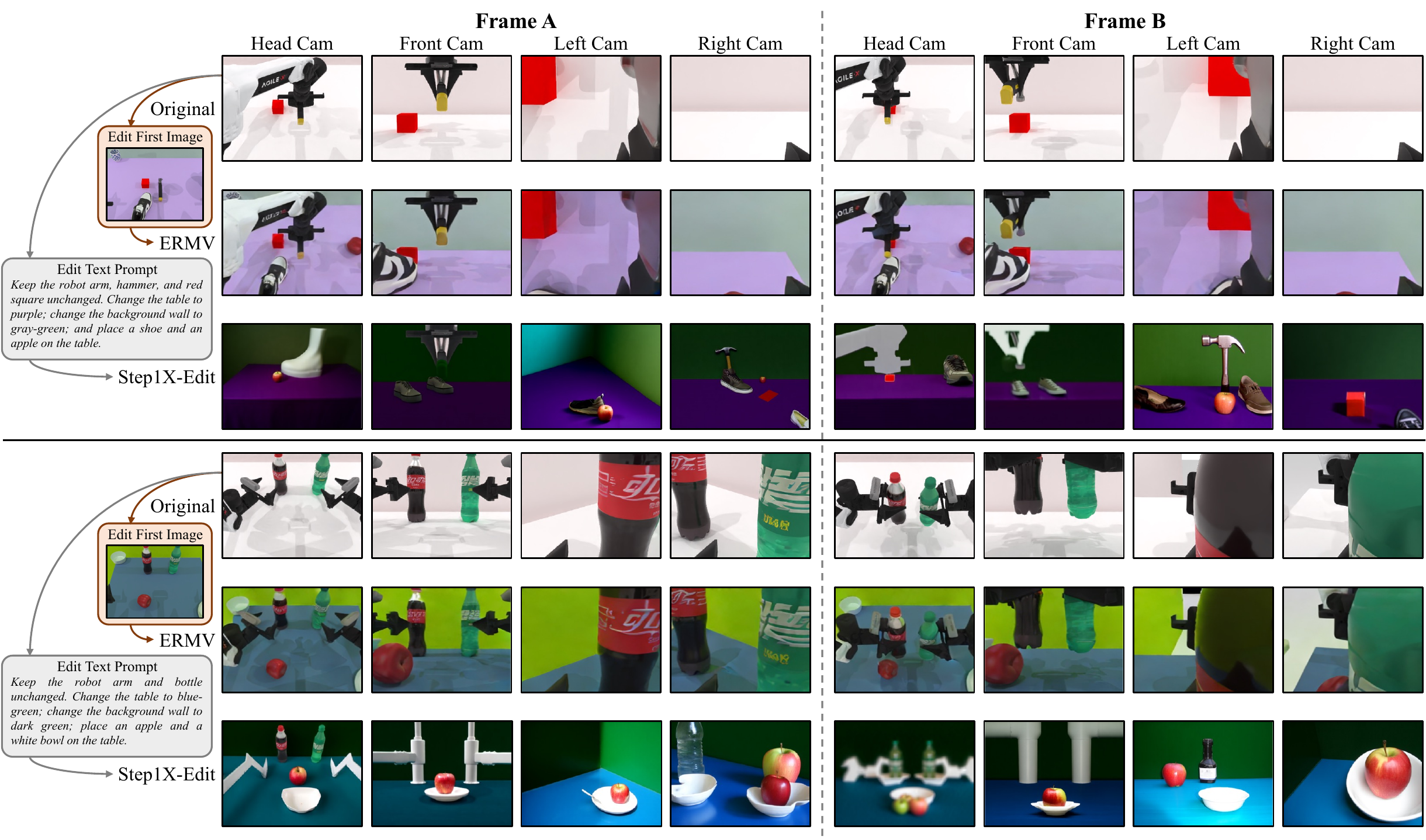}
    \vspace{-10pt}
   \caption{\textbf{Qualitative comparison} of editing 4D multi-view sequential images in \textbf{simulation environments}. ERMV is guided by the edited first frame head image. Step1X-Edit is guided by a text prompt.}
   \vspace{-8pt}
   \label{fig:sim_result}
\end{figure*}

\section{Experiments}
\label{Experiments}
This section comprehensively evaluates the performance of ERMV on the task of multi-view temporal image editing for robot manipulation. We begin by introducing the experimental setup. Subsequently, through a series of experiments in a simulated environment, we quantitatively assess the effectiveness of ERMV as a data augmentation technique and its ability to enhance the performance of downstream embodied agent policies. We then validate the editing quality of ERMV on a public real-world dataset. Furthermore, we deploy and test ERMV on a physical robot platform to examine its practical applicability in the physical world. Finally, through detailed ablation studies, we analyze the contributions of the key components of our model.

\subsection{Implementation Details}
\label{settings}
ERMV employs the U-Net backbone of Stable Diffusion 2.1 \cite{rombach2022high}. The model is trained with a batch size of 4. We use the AdamW optimizer with a constant learning rate of 1e-5. All models are implemented in PyTorch and are trained and evaluated on a single NVIDIA RTX 4090 GPU. To balance generation quality and computational efficiency, we adopt the SST sampling strategy: the historical context window randomly samples images from 4 views across the past 8 frames, while the future action window samples images from 6 views across the future 8 frames. In the feedback intervention mechanism, we utilize Qwen2.5-VL \cite{bai2025qwen2} as the Multimodal Large Language Model (MLLM) to assess and guide the generation process.

\begin{table}[t]\small 
\centering
\caption{\textbf{Quantitative results} of different methods for editing robotic 4D sequences in simulation environments.}

\vspace{-7pt}
\setlength{\tabcolsep}{2.3mm}
\renewcommand\arraystretch{1.0}
\begin{tabular}{l||ccccc}
\toprule
Method      & SSIM ↓ & PSNR ↑    & LPIPS ↓  \\ \hline
Step1X-Edit \cite{liu2025step1x} & 0.1916 & 6.31 & 0.6461 \\
ERMV        & \textbf{0.8334} & \textbf{24.17} & \textbf{0.1043} \\ \bottomrule
\end{tabular}
\label{tab:sim_table}
\vspace{-15pt}
\end{table}

\subsection{Simulation Experiments}
\label{sim_exp}
We conduct experiments on the dual-arm simulation platform RoboTwin \cite{mu2025robotwin}, which provides a suite of standardized robot manipulation tasks. For all tasks, we collect 4D trajectory data $\mathcal{T} = (\mathbf{X}_t, \mathbf{a}_t), t=1 \cdots T$, including multi-view images, robotic and camera states for model training. We asked ERMV to edit the collected data to augment the training data. In addition, the SOTA single-image editing method Step1X-Edit \cite{liu2025step1x} is also used as a comparison.

\begin{table*}[t]\small 
\centering
\caption{\textbf{Quantitative results} of training VLA models on data augmented by editing methods and tested in the \textbf{original scenes}. The Success Rate (SR) ↑ of 100 trials is used as the metric.}

\vspace{-7pt}
\setlength{\tabcolsep}{3.5mm}
\renewcommand\arraystretch{1.0}
\begin{tabular}{l||cccccc}
\toprule
\multirow{3}{*}{Tasks} & \multicolumn{6}{c}{Methods}                         \\ \cline{2-7} 
                       & \begin{tabular}[c]{@{}c@{}}RDT \cite{liu2024rdt}\\ (Baseline)\end{tabular} & \begin{tabular}[c]{@{}c@{}}RDT+\\ Step1X-Edit \cite{liu2025step1x}\end{tabular} & \multicolumn{1}{c|}{\begin{tabular}[c]{@{}c@{}}RDT+\\ ERMV\end{tabular}} & \begin{tabular}[c]{@{}c@{}}DP \cite{chi2023diffusion}\\ (Baseline)\end{tabular} & \begin{tabular}[c]{@{}c@{}}DP+\\ Step1X-Edit \cite{liu2025step1x}\end{tabular} & \begin{tabular}[c]{@{}c@{}}DP+\\ ERMV\end{tabular} \\ \hline
block hammer beat      & 0.55 & 0.00 & \multicolumn{1}{c|}{\textbf{0.59}}    & 0.00 & 0.00 & 0.00      \\
block handover         & 0.77 & 0.01 & \multicolumn{1}{c|}{\textbf{0.88}}    & 0.75 & 0.00 & \textbf{0.79}   \\
bottle adjust          & 0.25 & 0.00 & \multicolumn{1}{c|}{\textbf{0.52}}    & 0.35 & 0.00 & \textbf{0.44}   \\
container place        & 0.24 & 0.00 & \multicolumn{1}{c|}{\textbf{0.33}}    & 0.14 & 0.00 & \textbf{0.29}   \\
diverse bottles pick   & 0.11 & 0.00 & \multicolumn{1}{c|}{\textbf{0.13}}    & 0.12 & 0.00 & \textbf{0.13}   \\
dual bottles pick easy & 0.56 & 0.00 & \multicolumn{1}{c|}{\textbf{0.72}}    & 0.85 & 0.00 & \textbf{0.86}   \\
dual bottles pick hard & 0.37 & 0.00 & \multicolumn{1}{c|}{\textbf{0.41}}    & 0.59 & 0.00 & \textbf{0.61}   \\
empty cup place        & 0.13 & 0.00 & \multicolumn{1}{c|}{\textbf{0.19}}    & 0.87 & 0.00 & \textbf{0.89}   \\
pick apple messy       & 0.28 & 0.00 & \multicolumn{1}{c|}{\textbf{0.32}}    & 0.29 & 0.00 & \textbf{0.31}   \\
put apple cabinet      & 0.72 & 0.01 & \multicolumn{1}{c|}{\textbf{0.78}}    & 0.08 & 0.00 & \textbf{0.15}   \\
shoe place             & 0.21 & 0.00 & \multicolumn{1}{c|}{\textbf{0.23}}    & 0.33 & 0.00 & \textbf{0.36}   \\
tool adjust            & 0.55 & 0.00 & \multicolumn{1}{c|}{\textbf{0.66}}    & 0.04 & 0.00 & \textbf{0.07}   \\ \hline
Average            & 0.40 & 0.00 & \multicolumn{1}{c|}{\textbf{0.48}}    & 0.37 & 0.00 & \textbf{0.41}   \\ \bottomrule
\end{tabular}
\label{tab:sim_sr}
\vspace{-12pt}
\end{table*}

We first compare the quantitative results of the editing effects of different methods in the simulation environment, as shown in TABLE \ref{tab:sim_table}. As with the metrics used in other image editing methods \cite{lin2024schedule}, SSIM (Structural Similarity Index), PSNR (Peak Signal-to-Noise Ratio), LPIPS (Learned Perceptual Image Patch Similarity) are used as evaluation metrics. The results show that the editing results of ERMV are substantially ahead of the single-frame editing method Step1X-Edit.This is due to the excellent spatio-temporal consistency that ERMV maintains through EMA-Attn. Furthermore, the qualitative comparison results are shown in Fig. \ref{fig:sim_result}, ERMV achieves high fidelity editing effects. In particular, the shadow on the table and the light refraction on the surface of the bottle are edited accurately. This is attributed to the visual guidance condition accurately representing the desired effect in detail. The entire edited 4D sequence accurately responds to the scene transformation effects guided by the first edited image. In contrast, even the SOTA single-image editing method, Step1X-Edit, is guided by the text prompt, which makes it difficult to accurately express the desired editing effect and even completely destroys the semantics of the original image. Furthermore, the consistency between multiple views of the same frame after ERMV editing is accurately maintained. This is because the epipolar motion-aware attention module proposed by ERMV utilizes multi-view geometric constraints to ensure highly consistent static backgrounds from different viewpoints. Meanwhile, the SST module combined with motion injection effectively keeps the motion of the manipulated object and the robotic arm coherent with the history frames, ensuring smooth spatio-temporal consistency. On the contrary, Step1X-Edit edits completely different content even with the same text prompts because there is no mechanism to maintain temporal consistency.

To quantify the effectiveness of data generated by ERMV, we use it as a data augmentation method to train downstream embodied agent policies. We select RDT \cite{liu2024rdt} and Diffusion Policy (DP) \cite{chi2023diffusion} as the policy models. There are three training configurations: ``Baseline", where policy models are trained only on the original collected simulation data; ``+Step1X-Edit" replaces 80\% of the original data with Step1X-Edit augmented data for training; ``+ERMV" also replaces 80\% of the original data with ERMV augmented data for training. We then evaluate the average Success Rate (SR) of the policy models trained under different configurations in the standard test task of RoboTwin.

As presented in TABLE \ref{tab:sim_sr}, the model augmented with ERMV-generated data (``+ERMV") shows a significant improvement in success rate over the baseline in RDT (AVG: 0.40 vs. 0.48) and DP (AVG: 0.37 vs. 0.41). This is due to the fact that the baseline model was only trained on single simple scenes. Whereas the ERMV-enhanced data contained a variety of complex scenes. This result confirms the validity of the data augmented by ERMV, stemming from ERMV's strong ability to maintain spatio-temporal consistency. The SST module, in particular, ensures the continuity of the manipulation images across the entire time-series range, thus providing high-quality and physically consistent training signals for the policy models. The Step1X-Edit edited data, however, leads to a serious degradation of the performance of the VLA models, as it severely destroys the semantics of the original images.

\begin{figure*}[h]
  \centering
   \includegraphics[width=0.90\linewidth]{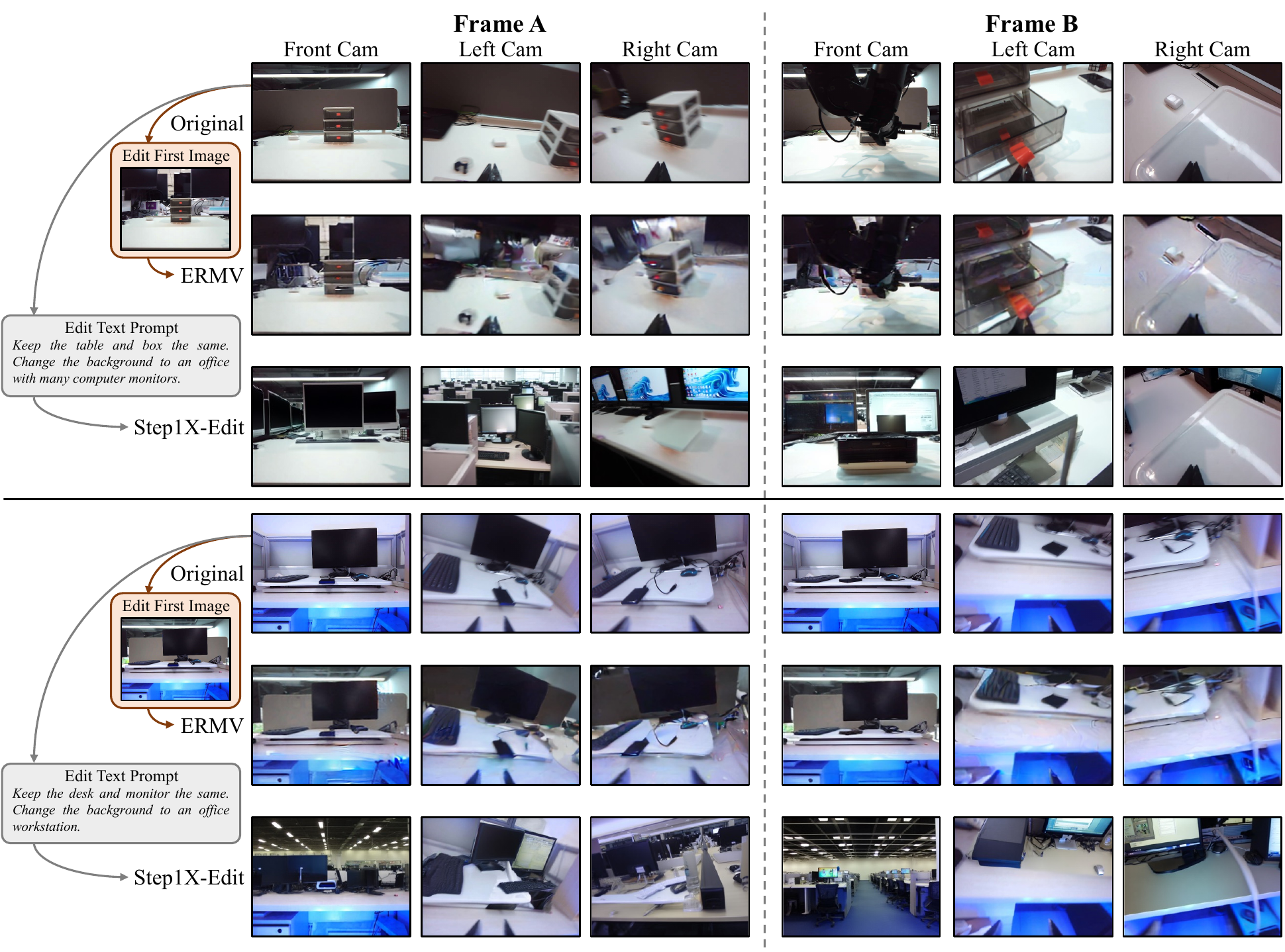}
    \vspace{-10pt}
   \caption{\textbf{Qualitative comparison} of editing 4D multi-view sequential images in \textbf{real-world environments}. ERMV is guided by the edited first frame front image. Step1X-Edit is guided by the text prompt.}
   \vspace{-8pt}
   \label{fig:real_result}
\end{figure*}

\begin{table*}[t]\small 
\centering
\caption{{\textbf{Quantitative results} of training VLA models on data augmented by editing methods and tested in the \textbf{unseen cluttered scenes}. The Success Rate (SR) ↑ of 100 trials is used as the metric.}}
\vspace{-7pt}
\setlength{\tabcolsep}{3.5mm}
\renewcommand\arraystretch{1.0}
\begin{tabular}{l||cccccc}
\toprule
\multirow{3}{*}{Tasks} & \multicolumn{6}{c}{Methods}                         \\ \cline{2-7} 
                       & \begin{tabular}[c]{@{}c@{}}RDT \cite{liu2024rdt}\\ (Baseline)\end{tabular} & \begin{tabular}[c]{@{}c@{}}RDT+\\ Step1X-Edit \cite{liu2025step1x}\end{tabular} & \multicolumn{1}{c|}{\begin{tabular}[c]{@{}c@{}}RDT+\\ ERMV\end{tabular}} & \begin{tabular}[c]{@{}c@{}}DP \cite{chi2023diffusion}\\ (Baseline)\end{tabular} & \begin{tabular}[c]{@{}c@{}}DP+\\ Step1X-Edit \cite{liu2025step1x}\end{tabular} & \begin{tabular}[c]{@{}c@{}}DP+\\ ERMV\end{tabular} \\ \hline
block hammer beat      & 0.10    & 0.00                 & \multicolumn{1}{c|}{\textbf{0.22}}    & 0.00     & 0.00              & 0.00                          \\
block handover         & 0.16    & 0.00                & \multicolumn{1}{c|}{\textbf{0.58}}    & 0.13      & 0.00             & \textbf{0.52}                       \\
bottle adjust          & 0.20    & 0.00                 & \multicolumn{1}{c|}{\textbf{0.56}}    & 0.24     & 0.00              & \textbf{0.48}                       \\
container place        & 0.05    & 0.00                & \multicolumn{1}{c|}{\textbf{0.38}}    & 0.02      & 0.00             & \textbf{0.33}                       \\
diverse bottles pick   & 0.07    & 0.00                & \multicolumn{1}{c|}{\textbf{0.15}}    & 0.12      & 0.00             & \textbf{0.13}                       \\
dual bottles pick easy & 0.31    & 0.00                & \multicolumn{1}{c|}{\textbf{0.70}}     & 0.33     & 0.00              & \textbf{0.68}                       \\
dual bottles pick hard & 0.14    & 0.00                & \multicolumn{1}{c|}{\textbf{0.21}}    & 0.24      & 0.00             & \textbf{0.37}                       \\
empty cup place        & 0.10    & 0.00                 & \multicolumn{1}{c|}{\textbf{0.16}}    & 0.31     & 0.00              & \textbf{0.61}                       \\
pick apple messy       & 0.17    & 0.00                & \multicolumn{1}{c|}{\textbf{0.19}}    & 0.19      & 0.00             & \textbf{0.23}                       \\
put apple cabinet      & 0.39    & 0.00                & \multicolumn{1}{c|}{\textbf{0.44}}    & 0.01      & 0.00             & \textbf{0.13}                       \\
shoe place             & 0.11    & 0.00                & \multicolumn{1}{c|}{\textbf{0.22}}    & 0.14      & 0.00             & \textbf{0.31}                       \\
tool adjust            & 0.43    & 0.00                & \multicolumn{1}{c|}{\textbf{0.58}}    & 0.01      & 0.00             & \textbf{0.09}                       \\ \hline
Average            & 0.19    & 0.00                & \multicolumn{1}{c|}{\textbf{0.37}}    & 0.15      & 0.00             & \textbf{0.32}                       \\ \bottomrule
\end{tabular}
\label{tab:sim_rand}
\vspace{-10pt}
\end{table*}

To comprehensively assess the generalization ability of the augmented policy models, we create more challenging ``clutter scene" based on the original test task of RoboTwin. To this end, we introduced random distracting objects into the environment and randomized the texture and background of the table, while keeping the core manipulated objects unchanged.

Notably, in the zero-shot generalization test of the novel ``unseen clutter scene", the performance of the baseline model dropped dramatically. This is because the baseline model is trained on a very singular scene. By contrast, the model trained with ERMV augmented data exhibits superior robustness and generalization ability, with the success rates far exceeding those of the baseline model in RDT (AVG: 0.19 vs. 0.37) and DP (AVG: 0.15 vs. 0.32). This result provides strong evidence that ERMV is a powerful data augmentation engine that can significantly enhance the generalization capabilities of downstream policies by creating diverse, high-quality out-of-domain training data. Through controlled editing of scene elements, ERMV can easily augment existing high-quality data. This enhanced robustness directly mitigates the challenges associated with collecting large-scale and diverse data.

\subsection{Real-World Experiments}
\subsubsection{Real-World Dataset Experiments}
To assess the editing capabilities and long-horizon stability of ERMV in real-world scenarios, we conduct experiments on the public dual-arm manipulation dataset, RDT-ft-data \cite{liu2024rdt}.

As illustrated in Fig. \ref{fig:real_result}, ERMV can successfully edit real-world robot manipulation sequences, such as replacing the background and table environment for the same grasping action. Notably, the model accurately preserves the morphology and motion of the core manipulated object, like the grasped box, and the robot arm during editing. This is primarily attributed to our EMA-Attn mechanism, which models multi-view geometric relationships to effectively distinguish between the dynamic foreground and static background, thereby enabling precise preservation of the manipulated objects. Furthermore, the edited images can even accurately reproduce motion blur effects caused by camera movement or rapid robot arm motions. This demonstrates that the multi-layer injection of motion information successfully captures and renders these delicate dynamic features in the robot state. While Step1X-Edit is able to edit the original image to the corresponding style based on text prompts, it not only destroys the semantics of a single frame, but the temporal changes are also inconsistent.

\subsubsection{Real Robot Experiments}
\begin{figure}[h]
  \centering
   \includegraphics[width=0.90\linewidth]{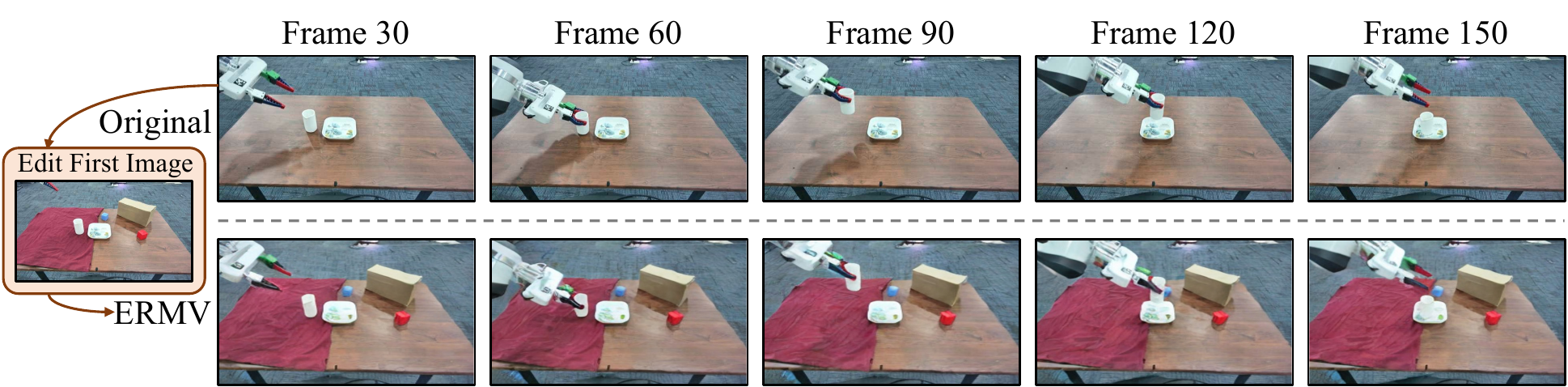}
    \vspace{-10pt}
   \caption{\textbf{Qualitative comparison} of editing 4D multi-view sequential images in \textbf{real robot experiments}. ERMV is guided by the edited first frame front image.}
   \vspace{-8pt}
   \label{fig:real_robot_result}
\end{figure}

\begin{figure*}[t]
  \centering
   \includegraphics[width=0.90\linewidth]{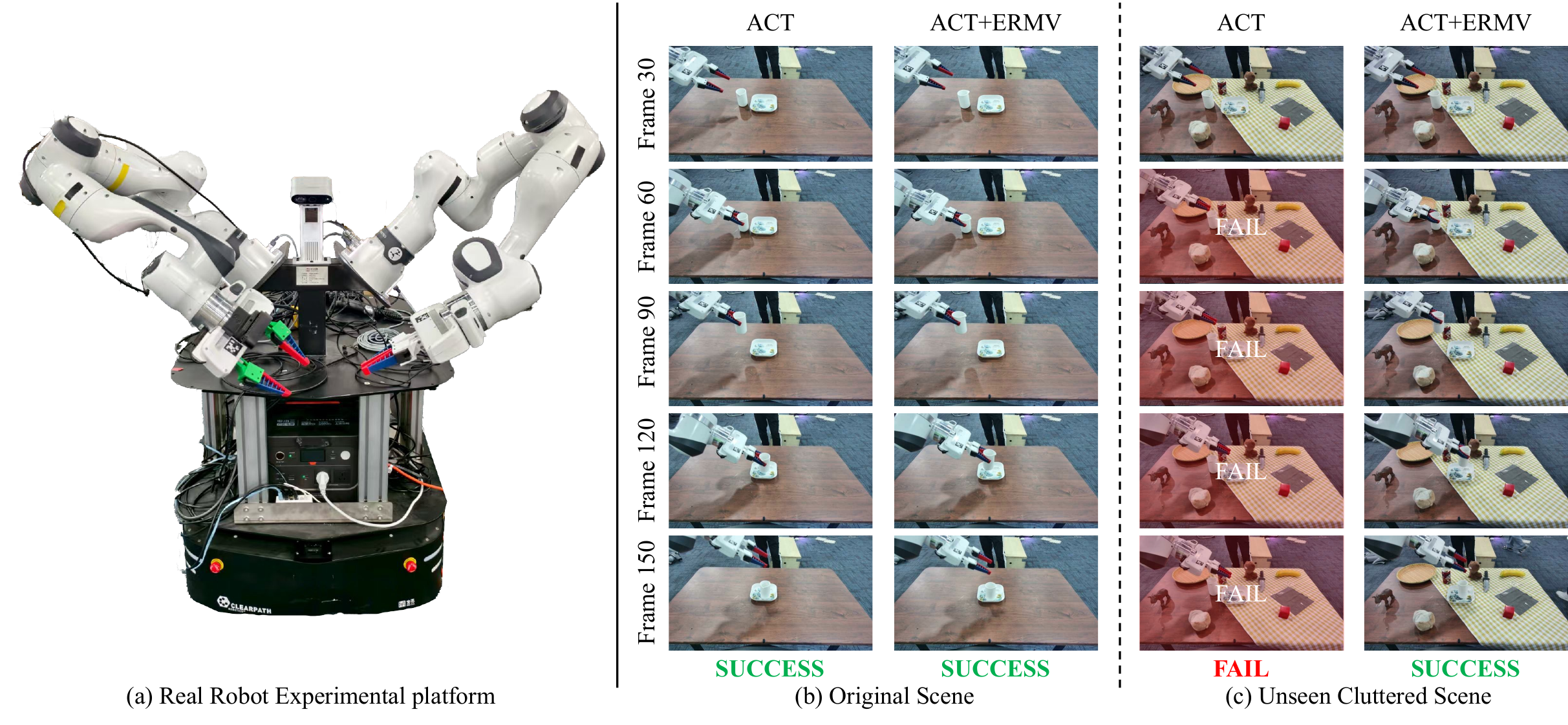}
    \vspace{-15pt}
   \caption{\textbf{Experimental results for real robot.} (a) Dual-arm robot platform for real robot experiments; (b) In the original scene, both ACT and ACT+ERMV can complete the task; (c) In an unseen cluttered scene, ACT cannot accurately pick up the target object. However, after ERMV augmentation training, the robustness of ACT+ERMV is improved, and it still accurately completes the task.}
   \vspace{-8pt}
   \label{fig:realrobot}
\end{figure*}

We further conducted physical experiments on a customized dual-arm robot platform consisting of two Franka Emika Panda robotic arms, as shown in Fig. \ref{fig:realrobot} (a). We first collected manipulation data for several pick-and-place tasks, on which we trained an Action Chunking with Transformers (ACT) \cite{zhao2023learning} policy model as a baseline. Subsequently, as shown in Fig. \ref{fig:real_robot_result}, we edited these data with ERMV to augment the training set and retrained the ACT model.

\begin{table}[t]\scriptsize 
\centering
\caption{{\textbf{Quantitative results} of training ACT on data augmented by editing methods and tested in the \textbf{real robot environment}. The Success Rate (SR) ↑ of 100 trials is used as the metric.}}
\vspace{-7pt}
\setlength{\tabcolsep}{1.5mm}
\renewcommand\arraystretch{1.1}
\begin{tabular}{l||cccc}
\toprule
\multicolumn{1}{l||}{\multirow{3}{*}{Tasks}} & \multicolumn{2}{c|}{Original Scene} & \multicolumn{2}{c}{Unseen Cluttered Scene} \\ \cline{2-5} 
\multicolumn{1}{c||}{}                       & \begin{tabular}[c]{@{}c@{}}ACT \cite{zhao2023learning}\\ (Baseline)\end{tabular} & \multicolumn{1}{c|}{ACT+ERMV}      & \begin{tabular}[c]{@{}c@{}}ACT \cite{zhao2023learning}\\ (Baseline)\end{tabular} & ACT+ERMV      \\ \hline
place the cup left   & 0.58   & \multicolumn{1}{c|}{\textbf{0.95}}     & 0.03    & \textbf{0.90}      \\
place the cup right   & 0.56    & \multicolumn{1}{c|}{\textbf{0.93}}    & 0.01   & \textbf{0.91}       \\
Average                             & 0.52   & \multicolumn{1}{c|}{\textbf{0.91}}    & 0.02    & \textbf{0.89}      \\ \bottomrule
\end{tabular}
\label{tab:real_robot_sr}
\vspace{-10pt}
\end{table}

As shown in Fig. \ref{fig:realrobot} (b), we first conducted tests in a simple, original scene. Both ACT and the augmented ACT+ERMV successfully completed the task. However, in a cluttered and unseen scene (Fig. \ref{fig:realrobot} (c)), the baseline ACT fails to grasp the object correctly due to excessive disturbances. Since ERMV augments data by editing previously collected data, the trained ACT+ERMV significantly enhances robustness. ACT+ERMV can still successfully complete the task in unseen cluttered scenes. 

The quantitative experimental results in TABLE \ref{tab:real_robot_sr} demonstrate that in the original scenes, the average success rate of ACT+ERMV after augmentation training increased from 0.52 to 0.91. This indicates that the augmented data from ERMV can enhance the stability of the downstream VLA model. In unseen cluttered scenes, the effect of ERMV is even more pronounced. The average success rate of ACT+ERMV remains at 0.89, whereas the success rate of the baseline ACT is only 0.02. This demonstrates that the robustness of ACT+ERMV has been greatly enhanced. This effect is attributed to ERMV accurately editing the collected data. This high-quality and diverse augmented data enables the downstream policy model to learn features that are more robust to visual changes in the real world, thereby effectively improving its performance in the physical world. This experimental result also confirms the conclusions drawn in Section \ref{sim_exp} in the simulation environment.

\begin{figure}[h]
  \centering
   \includegraphics[width=0.99\linewidth]{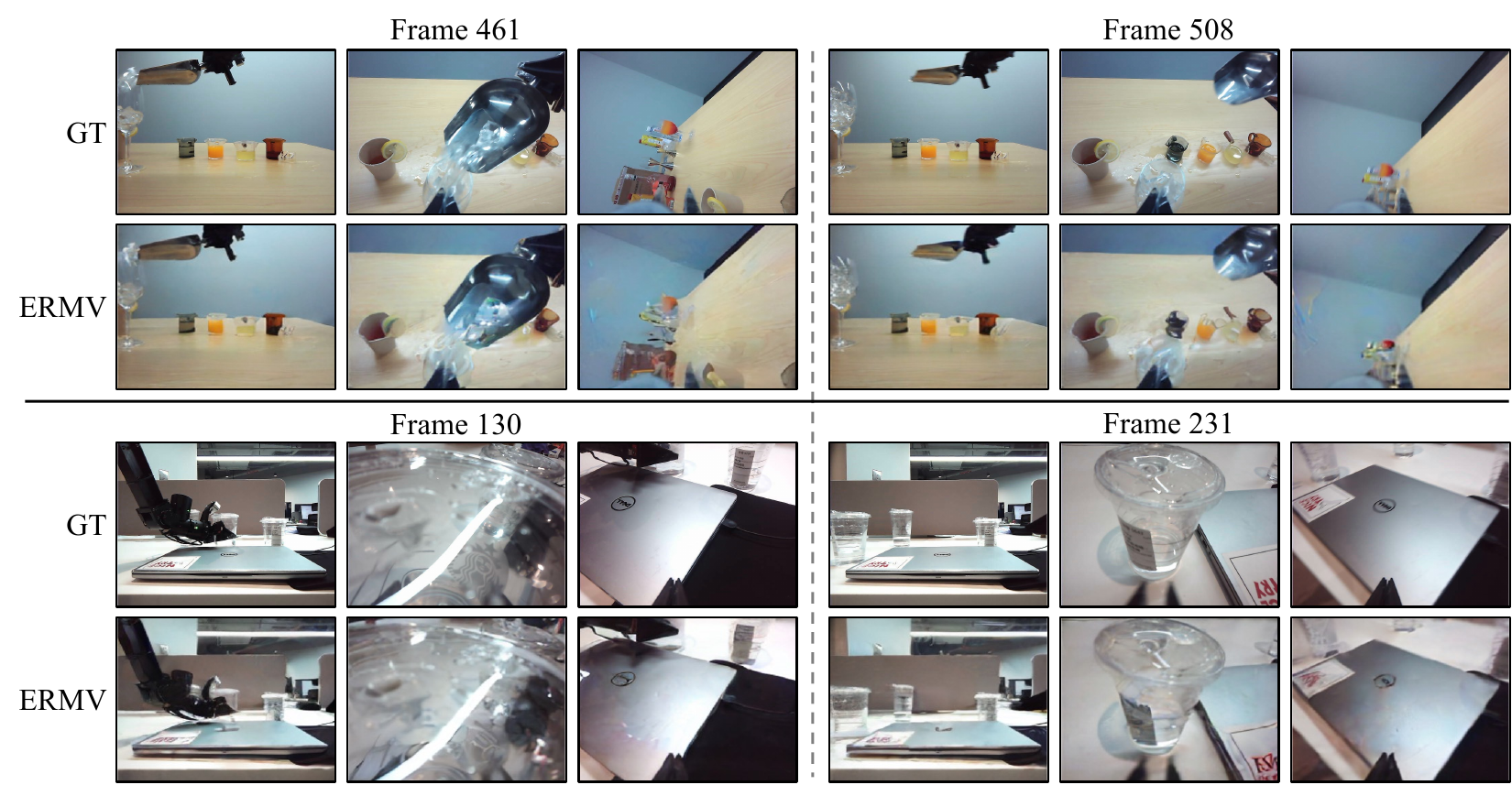}
    \vspace{-10pt}
   \caption{\textbf{Results of ERMV as a world model.} Guided by the first frame image and robot actions, ERMV can be used as a world model to generate complete sequences of images.}
   \vspace{-8pt}
   \label{fig:policy_val}
\end{figure}

\begin{figure}[t]
  \centering
   \includegraphics[width=0.99\linewidth]{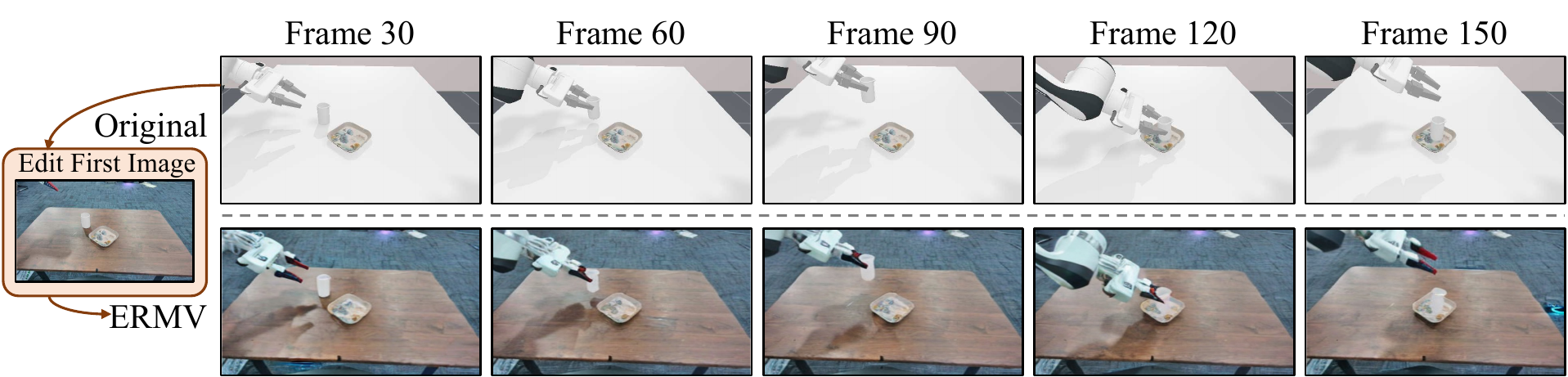}
    \vspace{-10pt}
   \caption{\textbf{ERMV edits virtual images as real scene style images.} Conditioned on an edited image and simulation actions, ERMV can convert the data collected in the simulation environment to real-world style data. This can make up for the sim-to-real gap and quickly expand real data by utilizing the convenience of collecting data in the simulation environment.}
   \vspace{-8pt}
   \label{fig:sim2real}
\end{figure}

\begin{figure}[t]
  \centering
   \includegraphics[width=0.99\linewidth]{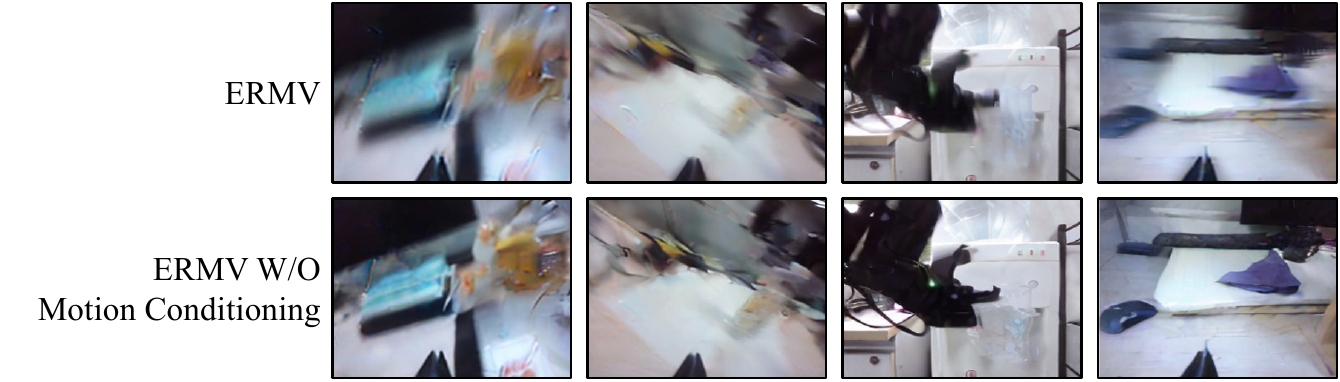}
    \vspace{-10pt}
   \caption{\textbf{The effect of Motion Conditioning in ERMV.} Benefiting from the multi-layer injection of motion information, ERMV can effectively edit images with motion blur.}
   \vspace{-8pt}
   \label{fig:ablation_motion}
\end{figure}

\begin{figure*}[t]
  \centering
   \includegraphics[width=0.99\linewidth]{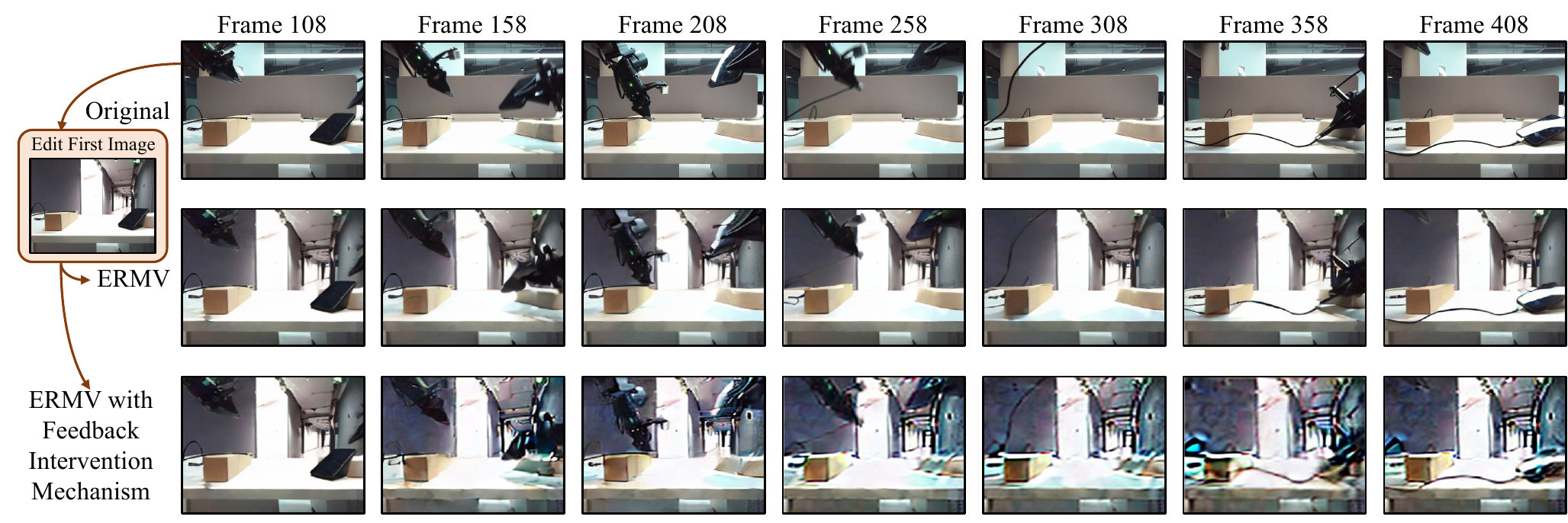}
    \vspace{-10pt}
   \caption{\textbf{Comparison of correcting cumulative errors using feedback intervention mechanism.} As the editing sequence increases, ERMV can effectively alleviate the problem of image degradation caused by cumulative error. In contrast, the image degradation is severe without feedback intervention.}
   \vspace{-8pt}
   \label{fig:long_horizon}
\end{figure*}

\subsection{Generation Capabilities}
An emerging application of ERMV is to serve as a world model for low-cost, high-efficiency validation of embodied agents without physical interaction. Moreover, editing simulated images into realistic scenes is also a novel application, which can compensate for the sim-to-real gap. We verify this through two experiments.

\textbf{World Model for Policy Validation.} When conditioned on a single initial raw frame and a sequence of actions $\mathbf{a}_t$ from VLA models, ERMV can be used as a world model to predictively generate corresponding multi-view spatio-temporal image sequences. As shown in Fig. \ref{fig:policy_val}, the generated interactive sequences are highly consistent with the Ground Truth (GT) images. Such accurate predictions are primarily due to our robot and camera state injection mechanism, which ensures the generation process strictly adheres to the input action commands. This demonstrates that ERMV can act as a reliable and deterministic world model for rehearsing and validating robot policies, thereby significantly accelerating the policy iteration cycle, avoiding risky physical trials at immature stages and eliminating the need to build high-fidelity simulation environments.

\textbf{Bridging the Sim-to-Real gap.} We conducted an experiment to explore the potential of ERMV in bridging the simulation-to-reality gap. ERMV first edits the initial frame of a simulation trajectory with a real-world visual style. Then, using this as the visual condition along with the original robot action sequence from the simulation, ERMV edits a complete ``pseudo-real" 4D multi-view trajectory that is realistic in appearance and physically consistent in motion. As shown in Fig. \ref{fig:sim2real}, the generated data successfully fuses the textures and lighting of a real scene with coherent physical actions. We use this ``pseudo-real" data to train ACT and evaluate it on a real robot. The Fig. \ref{fig:sim2real} shows that this ACT can accomplish tasks directly in real scenarios, which is a strong demonstration of the potential of ERMV in alleviating the scarcity of real data and bridging the gap between simulation and reality.

\subsection{Ablation Study}
\label{ablation}
To validate the effectiveness of each key component in ERMV, we conducted a comprehensive set of ablation studies.

\textbf{Effect of Motion Conditioning.} We removed the Motion Dynamics Conditioning and EMA-Attn modules. As depicted in Fig. \ref{fig:ablation_motion}, the model loses the ability to accurately capture motion information, thus failing to generate images with realistic motion blur effects. Although the generated images are visually ``sharper”, they lack the physical characteristics captured by a real camera. This demonstrates that the multi-layer injection of motion information is able to effectively simulate the dynamics of both the camera and the robotic arm.

\begin{table}[t]\small 
\centering
\caption{\textbf{Comparison using sparse spatio-temporal module with ERMV on RDT in RoboTwin.} Dense indicates dense approach and sparse indicates SST approach. The Success Rate (SR) ↑ of 100 trials is used as the metric.}
\vspace{-7pt}
\setlength{\tabcolsep}{1.5mm}
\renewcommand\arraystretch{1.0}
\begin{tabular}{l||cc}
\toprule
\multirow{2}{*}{Tasks} & \multicolumn{2}{c}{Methods}                           \\ \cline{2-3} 
                       & \multicolumn{1}{c}{ERMV with dense} & \multicolumn{1}{c}{ERMV with sparse} \\ \hline
block hammer beat      & 0.18       & \textbf{0.22}    \\
block handover         & 0.55   & \textbf{0.58} \\
bottle adjust          & 0.54     & \textbf{0.56}     \\
container place        & 0.37       & \textbf{0.38}   \\
diverse bottles pick   & 0.09     & \textbf{0.15}    \\
dual bottles pick easy & 0.63 & \textbf{0.70}\\
dual bottles pick hard & 0.17     & \textbf{0.21}      \\
empty cup place        & 0.13    & \textbf{0.16}  \\
pick apple messy       & 0.16  & \textbf{0.19}   \\
put apple cabinet      & 0.36    & \textbf{0.44}    \\
shoe place             & 0.19     & \textbf{0.22}  \\
tool adjust            & 0.44    & \textbf{0.58}    \\ \hline
Average            & 0.32    & \textbf{0.37}     \\ \bottomrule
\end{tabular}
\label{tab:ablation_efficiency}
\vspace{-10pt}
\end{table}

\textbf{Efficiency of Sparse Spatio-Temporal Module.} We compare the performance of the sparse approach with the dense approach. TABLE \ref{tab:ablation_efficiency} shows that the SST module in ERMV achieves better performance. Because the sparse approach can set a larger working window than dense sampling with the same GPU memory. In this way, the sparse approach allows better extraction of history information and maintains consistency over long time series.

In addition, when the same working window is fixed, the sparse approach can significantly reduce the need for GPU memory by 50\%. This allows ERMV to be trained on consumer GPUs with small GPU memory, greatly improving the utility and scalability of the algorithm.

\textbf{Effect of Feedback Intervention Mechanism.} When processing long-horizon 4D data, conventional autoregressive models often suffer from semantic drift and detail blurring due to error accumulation. As shown in the comparative experiment in Fig. \ref{fig:long_horizon}, we disabled the feedback intervention strategy to evaluate its effect. The model without this strategy suffers from gradual quality degradation, exhibiting severe artifacts and semantic drift due to error accumulation. In contrast, the full ERMV model maintains a high-quality output throughout the sequence. This is credited to the feedback intervention mechanism, which performs self-assessment during inference to promptly detect and correct potential biases, thereby ensuring high-quality and consistent editing over long sequences.

\section{Discussion}
In this paper, we introduced the ERMV framework, with the primary objective of breaking the data bottleneck in robot imitation learning. Our research substantiates a critical thesis: the performance of Visual-Language-Action (VLA) models can be significantly enhanced by efficiently and consistently editing existing high-quality data. Beyond architectural innovations, ERMV pragmatically incorporates a feedback intervention mechanism. This ``MLLM review + expert correction" paradigm offers a practical intermediate path toward building trustworthy AI systems. It is not only a technical tool to ensure data quality, but also an effective strategy to align the behavior of AI systems with high-level task goals, such as maintaining the physical realism of a robotic arm.

Furthermore, the powerful editing capabilities of ERMV open up a new paradigm for studying the generalization and robustness of robotic policies. By changing the background, lighting, and even object layout of a task scene, we can cost-effectively and massively edit existing high-quality data into test environments that are difficult to construct in the real world. This allows researchers to systematically expand the types of data available for robotics policies without investing substantial resources in building complex experimental scenes or high-fidelity simulation environments. ERMV can even edit sequences to obtain hazardous data that would be difficult to collect from real robots, such as pre-collision 4D robotic images. Similarly, its ability to act as a world model to generate continuous 4D data from a single frame and actions provides a safe and efficient offline evaluation solution for robot motion planning, effectively reducing the need for high-stakes testing on physical hardware.

Despite these encouraging results, we also recognize the limitations of ERMV.
The current framework of ERMV does not introduce data such as depth images, 3D Gaussian splatting, etc., which have rich 3D structural information. This is because the editing of these data is more complex than editing single-frame images. However, it is clear that the addition of these data can significantly improve the effectiveness of 4D data editing. We will explore how to introduce more 3D information to enhance performance in the future. Moreover, to fully automate the data editing pipeline, we will explore using advanced semantic segmentation or object detection techniques to replace parts of the current manual annotation and intervention processes, thus further improving the efficiency and scalability of the ERMV framework.

Finally, the foundational principles applied in ERMV, such as SST module and motion-aware attention, provide inspiration for other dynamic video generation fields. They may be available for broader applications.

\vspace{-6pt}
\section{Conclusion}
\vspace{-4pt}
In this paper, we alleviate a critical data bottleneck in robotic imitation learning by introducing ERMV, a novel framework for augmenting 4D multi-view sequential data. Guided by detailed single-frame image editing, ERMV efficiently and accurately controls the editing goal of the entire sequence. With the sparse spatio-temporal module, ERMV can maximize the working window with limited hardware. Furthermore, the epipolar motion-aware attention ensures multi-view consistency and motion blur restoration through geometric guidance. Moreover, the feedback intervention strategy effectively mitigates the error accumulation and improves the quality of autoregressive editing. Our extensive experiments demonstrate that data augmented by ERMV can significantly improve the performance and robustness of VLA models. In addition, ERMV can be used not only as a strategy evaluation tool, but also to bridge the gap between simulation and reality.

\bibliographystyle{IEEEtran}
\bibliography{main}












\vfill

\end{document}